\documentclass[sigconf]{acmart}
\renewcommand\footnotetextcopyrightpermission[1]{} 
\usepackage{amsmath, bm}
\usepackage{multirow}
\usepackage{subfigure}
\usepackage{algorithm}  
\usepackage{algorithmic} 
\usepackage{balance}
\usepackage{soul}
\usepackage{adjustbox}
\usepackage{array}
\usepackage{cleveref}

\newtheorem{problem}{Problem}
\newtheorem{proposition}{Proposition}[section]

\newcommand{\expect}[2]{\mathop{\mathbb{E}}\limits_{#1}\left [#2\right]}
\newcommand{\boldx}[0]{\boldsymbol{X}}
\newcommand{\bolds}[0]{\boldsymbol{S}}
\newcommand{\boldv}[0]{\boldsymbol{V}}
\newcommand{\boldb}[0]{\boldsymbol{\beta}}
\newcommand{\boldsigma}[0]{\boldsymbol{\Sigma}}
\newcommand{\boldtheta}[0]{\boldsymbol{\theta}}
\usepackage[normalem]{ulem}
\useunder{\uline}{\ul}{}

\AtBeginDocument{%
  }

\setcopyright{acmlicensed}
\copyrightyear{2018}
\acmYear{2018}
\acmDOI{XXXXXXX.XXXXXXX}
\acmISBN{978-1-4503-XXXX-X/2018/06}

\begin{document}

\title{Sample Weight Averaging for Stable Prediction}

\author{Han Yu}
\email{yuh21@mails.tsinghua.edu.cn}
\affiliation{%
  \institution{Tsinghua University}
  \city{}
  \country{}
}

\author{Yue He}
\email{hy865865@gmail.com}
\affiliation{%
  \institution{Tsinghua University}
  \city{}
  \country{}
}

\author{Renzhe Xu}
\email{xurenzhe@sufe.edu.cn}
\affiliation{%
  \institution{Shanghai University of Finance and Economics}
  \city{}
  \country{}
}

\author{Dongbai Li}
\email{ldb22@mails.tsinghua.edu.cn}
\affiliation{%
  \institution{Tsinghua University}
  \city{}
  \country{}
}

\author{Jiayin Zhang}
\email{jiayin-z24@mails.tsinghua.edu.cn}
\affiliation{%
  \institution{Tsinghua University}
  \city{}
  \country{}
}

\author{Wenchao Zou}
\email{wenchao.zou@siemens.com}
\affiliation{%
  \institution{Siemens China}
  \city{}
  \country{}
}

\author{Peng Cui}
\authornote{Corresponding Author}
\email{cuip@tsinghua.edu.cn}
\affiliation{%
  \institution{Tsinghua University}
  \city{}
  \country{}
}

\renewcommand{\shortauthors}{Yu et al.}

\begin{abstract}
The challenge of Out-of-Distribution (OOD) generalization poses a foundational concern for the application of machine learning algorithms to risk-sensitive areas. 
Inspired by traditional importance weighting and propensity weighting methods, 
prior approaches employ an independence-based sample reweighting procedure. They aim at decorrelating covariates to counteract the bias introduced by spurious correlations between unstable variables and the outcome, thus enhancing generalization and fulfilling stable prediction under covariate shift. 
Nonetheless, these methods are prone to experiencing an inflation of variance, primarily attributable to the reduced efficacy in utilizing training samples during the reweighting process. 
Existing remedies necessitate either environmental labels or substantially higher time costs along with additional assumptions and supervised information. 
To mitigate this issue, we propose SAmple Weight Averaging (SAWA), a simple yet efficacious strategy that can be universally integrated into various sample reweighting algorithms to decrease the variance and coefficient estimation error, thus boosting the covariate-shift generalization and achieving stable prediction across different environments. We prove its rationality and benefits theoretically. Experiments across synthetic datasets and real-world datasets consistently underscore its superiority against covariate shift. 
\end{abstract}


\maketitle

\section{Introduction}

Over the past decade, there have been substantial advancements in artificial intelligence, driven by the enhanced computational power and the strong generalization capabilities exhibited by neural networks~\cite{he2016deep, brown2020language}. 
However, the effectiveness of current algorithms relies heavily on the assumption that training data and test data are independent and identically distributed, known as the IID assumption. In cases where this assumption does not hold, a circumstance frequently encountered in real-world scenarios, their performance lacks theoretical and empirical guarantees. Models exhibit vulnerability to spurious correlations~\cite{sagawa2020investigation} or shortcuts~\cite{geirhos2020shortcut} and their performance may experience significant deterioration in the presence of distribution shifts. This limitation substantially impedes the application of machine learning in risk-sensitive areas like autonomous driving~\cite{muhammad2020deep} and law~\cite{surden2014machine}. 

Therefore, the problem of Out-of-Distribution (OOD) generalization has garnered significant attention within the realm of machine learning in recent years~\cite{shen2021towards}. Different from domain adaptation (DA)~\cite{ben2006analysis,long2015learning, chu2016selective} where test data information is known partially, OOD generalization addresses the challenging but more realistic setting where test data is totally unknown. 
Several lines of research have been dedicated to tackling this challenge. 
Both invariant learning~\cite{arjovsky2019invariant, krueger2021out} and domain generalization (DG)~\cite{shankar2018generalizing, rame2022fishr} aim to capture inherent invariance within training data to enhance the OOD generalization ability, while DG also involves techniques such as meta learning~\cite{li2018learning} and data augmentation~\cite{zhou2020domain}. Usually they require explicit environmental information during training.  
Another series of methods named distributionally robust optimization (DRO)~\cite{volpi2018generalizing, duchi2021learning} resolves around optimizing for the worst-case distribution to bolster OOD generalization performance, but they suffer from the over-pessimism problem~\cite{liu2022distributionally}. 

Recently, inspired by importance weighting~\citep{huang2006correcting,bickel2009discriminative} and propensity weighting~\citep{li2018balancing,lee2010improving}, 
stable learning algorithms~\citep{kuang2020stable, shen2020stable,zhang2021deep} employ independence-based sample reweighting to address covariate shift, which is a most common distribution shift~\citep{shen2021towards}. Without access to test data and explicit environment labels, they leverage a structural assumption that divides covariates into stable variables $\bolds$ and unstable variables $\boldv$, and try to mitigate correlations between $\bolds$ and $\boldv$ by reweighting to a distribution where covariates are independent. 
With theoretical guarantees~\citep{xu2021stable}, in such a way they can remove spurious correlations between $\boldv$ and the outcome $Y$ under the infinite-sample scenario. Thus they eliminate the bias of model parameter estimation and achieve stable prediction against agnostic covariate shift~\cite{cui2022stable}. 
Nevertheless, when it comes to the finite-sample scenario, despite the pronounced capability for debiasing, the operation of sample reweighting is naturally a less effective way of utilizing samples~\citep{kish1965survey}, i.e. diminishing the effective sample size~\citep{martino2017effective}. It leads to an increased variance of parameter estimation by all means, let alone the reweighting operation is too ambitious as it attempts to decorrelate all covariates without distinction. 
Thus they tend to fail to achieve a good bias-variance trade-off in situations featuring strong collinearity among covariates. While efforts have been devoted to tackling this issue, some require environment labels for covariate clustering~\citep{shen2020stable2}, and some assume low collinearity in unstable variables, 
need to take advantage of supervised information $Y$ and require much higher time cost due to the iterative procedure~\citep{yu2023stable}. 

It is well-known that bagging~\citep{breiman1996bagging}, a classic ensemble learning strategy, helps reduce the variance of the learned model. It constructs an ensemble of models trained on datasets bootstrapped from the original data, employing techniques such as averaging or majority voting. 
Inspired by this, we propose a universal and effective strategy of SAmple Weight Averaging (SAWA) to reduce the estimation variance and achieve a better bias-variance trade-off. 
It averages the sets of sample weights acquired by the identical weight learning procedure but with varying random initializations. 
Compared with previous remedies for strong collinearity and high variance, SAWA does not require environment labels or supervised information to conduct sample reweighting. It also lowers the time cost since it can be easily parallelized, unlike the iterative procedure in~\citet{yu2023stable}. 
We theoretically prove both the rationality and effectiveness of SAWA, and empirically demonstrate its usefulness and universality when incorporated into current independence-based sample reweighting schemes across various synthetic data settings. We also exhibit its practical usage of improving the covariate-shift generalization ability on real-world datasets. Additionally, we make further analyses to show the superiority of SAWA by comparing it with existing averaging or ensemble strategies in OOD research.

Our main contributions are listed as follows:
\begin{itemize}
    \item We introduce a strategy of averaging sample weights for stable prediction against covariate shift to mitigate concerns related to variance inflation.
    \item We provide theoretical evidence for the validity of the averaged sample weights and advantages of decreasing the error of weight learning and model parameter estimation. 
    \item We demonstrate the effectiveness and universality of the strategy through comprehensive experiments on both synthetic and real-world datasets.
\end{itemize}

\section{Related Work}

\textbf{Domain generalization} (DG) offers a solution for generalizing to out-of-distribution (OOD) scenarios where preliminary access to test data is lacking. Typically, these approaches necessitate the presence of multiple subpopulations, referred to as domains, within the training data. They learn representations that exhibit invariance across these subpopulations \citep{ghifary2015domain, li2018domain, shankar2018generalizing}. Additional methodologies include meta-learning \citep{li2018learning, balaji2018metareg, li2019episodic}, as well as the augmentation of source domain data \citep{qiao2020learning, zhou2020domain}. Recently, some methods based on parameter averaging have arisen \citep{cha2021swad, arpit2022ensemble} with impressive performance. 
Nevertheless, theoretical underpinning remains deficient within DG, and empirically, its efficacy heavily hinges on the availability of a plethora of training domains.

\textbf{Invariant learning} constitutes another branch of research that endeavors to algorithms with theoretical guarantees. In a manner akin to domain generalization, its objective is to capture the invariance amid diverse training environments, thereby augmenting the models' ability for generalization across unknown test distributions. Consequently, these approaches require either explicit environment labels \citep{arjovsky2019invariant, koyama2020out, krueger2021out} or the assumption of sufficient heterogeneity within the training data \citep{liu2021heterogeneous}. 

\textbf{Distributionally robust optimization} (DRO) focuses on optimizing the worst-case distribution. It chooses a perturbation ball around the original distribution and seeks the worst distribution in that ball for optimization. There are various ways to depict the distribution distance for the ball radius, including f-DRO \citep{duchi2021learning} and Wasserstein DRO \citep{sinha2018certifying}. It is worth noting that always focusing on the worst case naturally results in over-pessimism \citep{liu2022distributionally}.


\textbf{Stable learning} addresses covariate shift by introducing a structural assumption and applying sample reweighting. This approach typically does not require explicit inclusion of environmental information \citep{cui2022stable}. \citet{xu2021stable} conduct a comprehensive theoretical analysis of independence-based sample reweighting, serving as the theoretical foundation of stable learning. 
DWR \citep{kuang2020stable} and SRDO \citep{shen2020stable} employ global decorrelation of covariates via sample reweighting. \citet{shen2020stable2} adapt DWR into group-wise decorrelation to mitigate the variance inflation caused by global decorrelation, which is also addressed in~\citet{yu2023stable} through the design of an iterative framework incorporating sparsity constraint. 
Besides, \citet{zhang2021deep} extend stable learning empirically to deep learning.

\section{Method}

\subsection{Notations and Problem}
\label{sec:problem}

Throughout the paper, regular letters signify scalars, while bold letters denote vectors or matrices. 
For variables, uppercase letters are used. 
The input covariates are denoted as $\boldsymbol{X}\in \mathbb{R}^p$ with a dimension $p$, where $X_d$ represents the d$^{th}$ covariate. The outcome $Y\in \mathbb{R}$ corresponds to a scalar in the context of regression tasks. 
When referring to a dataset with a sample size $n$, lowercase letters are used: $\boldsymbol{x}\in \mathbb{R}^{n\times p}$ denotes the input data, also the design matrix, and $\boldsymbol{y}\in\mathbb{R}^n$ denotes the outcome data, where $(\boldsymbol{x}_i,y_i)$ denotes the i$^{th}$ sample. Training distribution is denoted as $P^{tr}(\boldsymbol{X}, Y)$ while test distribution is denoted as $P^{te}(\boldsymbol{X}, Y)$. 
The notation $\boldsymbol{A}\perp \boldsymbol{B}$ is employed to indicate the statistical independence between $\boldsymbol{A}$ and $\boldsymbol{B}$. 
Expectations are represented by $\mathbb{E}_{Q(\cdot)}[\cdot]$, where $Q$ can be chosen as $P^{tr}$, $P^{te}$ or any other proper distributions. 

We use $w(\boldsymbol{X})\in \mathcal{W}$ to denote the weighting function whose input is $\boldx$, formally defined below:
\begin{definition} [Weighting function] \label{def:weighting}
    Let $\mathcal{W}$ be the set of weighting functions that satisfy
    \begin{equation}
    \small
        \mathcal{W} = \left\{w: \mathcal{X} \rightarrow \mathbb{R}^{+} \mid \mathbb{E}_{P^{tr}(\boldsymbol{X})}[w(\boldsymbol{X})] = 1 \right\}.
    \end{equation}
    Then $\forall w \in \mathcal{W}$, the corresponding weighted distribution is $\tilde{P}_w(\boldsymbol{X}, Y) = w(\boldsymbol{X})P^{tr}(\boldsymbol{X}, Y)$. $\tilde{P}_w$ is well defined with the same support of $P^{tr}$.
\end{definition}
Regarding independence-based sample reweighting, $\mathcal{W}_{\perp}$ designates the subset of $\mathcal{W}$ where $\boldsymbol{X}$ are mutually independent in the weighted distribution $\tilde{P}_w$.
The parameters for weight learning are denoted as $\boldtheta\in \mathbb{R}^n$ while the parameters (coefficients) for linear regression are represented as $\boldsymbol{\beta}\in \mathbb{R}^p$ to establish a clear distinction. 
For a weight learning algorithm $\mathbb{A}$, it takes data $\boldsymbol{x}$ and parameter initialization $\boldtheta_0$ as input, and produces a weighting function parameterized by $\boldtheta$, expressed as $w_{\boldtheta}=\mathbb{A}(\boldsymbol{x},\boldtheta_0)\in \mathcal{W}$. 
In the scenarios of finite samples, $W_i$ is employed as an abbreviation for $w(\boldsymbol{x}_i)$, and $\boldsymbol{W}\in \mathbb{R}^n$ is utilized in place of $w$ without ambiguity.

Generally speaking, OOD generalization focuses on the setting where $P^{tr}(\boldsymbol{X}, Y)\neq P^{te}(\boldsymbol{X}, Y)$. For covariate-shift generalization, its definition is presented below.

\begin{problem}[Covariate-Shift Generalization]
    Given samples $\{(\boldsymbol{x}_i, y_i)\}_{i=1}^{n}$ drawn from training distribution $P^{tr}$, the goal of covariate-shift generalization is to learn a prediction model so that it performs stably on predicting $Y$ in agnostic test distribution where $P^{te}(\boldsymbol{X}, Y)=P^{te}(\boldsymbol{X})P^{tr}(Y|\boldsymbol{X})$.
\end{problem}
We can see $P^{te}$ differs from $P^{tr}$ in the shift of covariate distribution only, while keeping the conditional distribution fixed. 
To address this covariate-shift generalization problem, the concept of a minimal stable variable set is defined below.

\begin{definition} [Minimal stable variable set] 
    A minimal stable variable set of predicting $Y$ under training distribution $P^{tr}$ is any subset $\boldsymbol{S}$ of $\boldsymbol{X}$ satisfying the following equation, and none of its proper subsets satisfies it.
    \begin{equation}
    \small
        \mathbb{E}_{P^{tr}}[Y | \boldsymbol{S}] = \mathbb{E}_{P^{tr}}[Y | \boldsymbol{X}]. \label{eq:stable-set}
    \end{equation}
\end{definition}
\citet{xu2021stable} theoretically prove that under the assumption of strictly positive density, which is common in causal inference~\citep{imbens2015causal}, the minimal stable variable set $\bolds$ is equivalent to the minimal and optimal predictor under $P^{te}$. Thus we aim to capture the minimal stable variable set $\bolds$, i.e. stable variables, and get rid of unstable variables $\boldv=\boldx\backslash\bolds$.

\subsection{Independence-Based Sample Reweighting}

In different environments, correlations between $\bolds$ and $\boldv$ tend to vary, leading to the spurious correlations between $Y$ and $\boldv$ which could be easily learned by traditional machine learning algorithms. Thus previous stable learning methods try to decorrelate $\bolds$ and $\boldv$ via sample reweighting~\citep{kuang2020stable,shen2020stable,yu2023stable,xu2021stable}. Under the infinite-sample setting, \citet{xu2021stable} prove that by conducting weighted least squares (WLS) via $w\in \mathcal{W}_{\perp}$, where $\boldsymbol{X}$ are mutually independent after reweighting, coefficients on unstable variables $\boldsymbol{V}$ are zero almost surely even when the data generation function of $Y$ is nonlinear, serving as the foundation of independence-based sample reweighting against covariate shift. Two representative sample reweighting techniques are introduced below. 

\paragraph{DWR}~\citep{kuang2020stable} aims to remove pairwise linear correlations, i.e.
\begin{equation} \label{eq:DWR}
\small
    \hat{w}(\boldsymbol{X}) = \arg \min_{w(\boldsymbol{X})} \sum_{1\le i,j \le p, i\ne j}\left(Cov(X_i, X_j; w)\right)^2,
\end{equation}
where $Cov(X_i, X_j; w)$ represents the covariance of $X_i$ and $X_j$ after being reweighted to $\tilde{P}_{w}$. 
DWR is well fitted for the case where the data generation process is dominated by a linear function, since it focuses on linear decorrelation only. 

\paragraph{SRDO}~\citep{shen2020stable} conducts sample reweighting by density ratio estimation. 
It simulates the target distribution $\tilde{P}$ via random resampling on each covariate so that $\tilde{P}(X_1, X_2, \dots, X_p) = \prod_{i=1}^p P^{tr}(X_i)$. 
Then the weighting function can be estimated by
\begin{equation} \label{eq:SRDO}
\small
    \hat{w}(\boldsymbol{X}) = \frac{\tilde{P}(\boldsymbol{X})}{P^{tr}(\boldsymbol{X})} =  \frac{P^{tr}(X_1)P^{tr}(X_2)\dots P^{tr}(X_p)}{P^{tr}(X_1, X_2, \dots, X_p)}.
\end{equation}
To estimate such density ratio, with the help of MLP, we can learn a binary classifier to predict the probability of which distribution a sample belongs to, or employ LSIF loss for direct estimation~\citep{menon2016linking}. 
Compared with DWR, SRDO can not only decrease linear correlations among covariates, but weaken the nonlinear dependence. 

It is worth noting that both DWR and SRDO suffer from the over-reduced effective sample size and variance inflation~\citep{shen2020stable2} since they conduct global decorrelation between all covariates while strong correlations commonly exist inside stable variables. Therefore, they both require an enormous sample size to work. Although~\citet{yu2023stable} attempt to address this issue via an iterative framework, they require the utilization of the outcome $Y$ when learning sample weights and suffer from a high time cost considering the iterative process that cannot be parallelized.

\subsection{Sample Weight Averaging}

To mitigate the issue of low effective sample size and high variance in previous independence-based sample reweighting methods, we turn to bagging for inspiration. 
As a conventional ensemble learning strategy, bagging can decrease the estimation variance by averaging models trained on bootstrap-sampled data from the original dataset. 
Thus we consider designing a similar ensemble procedure. 
In order to generate diverse weighting functions, we note that in DWR, since the sample size, i.e. the number of parameters for sample weight learning, is much larger than the feature dimension, it could bear resemblance to the overparameterization characteristics of neural networks, e.g. one may anticipate the existence of multiple local minima when optimizing with gradient descent. The same is true for SRDO since the number of MLP parameters is much larger than the feature dimension. 
Consequently, we are likely to obtain diverse solutions even applying the same algorithm, as long as we vary elements of randomness like initialization. 
We theoretically substantiate this intuition in \Cref{prop:dwr} and \ref{prop:srdo}, and empirically confirm it in \Cref{fig:dist-comp} and \ref{fig:sim-comp} of \Cref{sec:synthetic}.

Thus we propose SAmple Weight Averaging (SAWA) to improve covariate-shift generalization ability of independence-based sample reweighting algorithms. It learns multiple sets of sample weights by varying the random initialization of parameters $\boldtheta$ in weight learning. 
For DWR, we adopt standard normal distribution to initialize sample weights. For SRDO, we use Xavier Glorot Initialization \citep{glorot2010understanding} for the MLP-structured weighting function. 
Then we directly average the set of sample weights to yield the ensemble result. The entire procedure is described in \Cref{alg:sawa}.

\begin{algorithm}[t]
\caption{SAmple Weight Averaging (SAWA)} \label{alg:sawa}
\begin{algorithmic}
    \STATE {\bfseries Input:} 
    \item Dataset $[\boldsymbol{x}, \boldsymbol{y}]$, where $\boldsymbol{x}\in \mathbb{R}^{n\times p}, \boldsymbol{y}\in \mathbb{R}^{n\times 1}$. 
    \item Weight learning algorithm $\mathbb{A}$.  
    \item Number of averaged sets of sample weights $K$. 
    
    \STATE {\bfseries Output:} Sample weights $\bar{\boldsymbol{W}}$. 
    \STATE Initialize $\tilde{\boldsymbol{W}}$ as an empty list.
    \FOR {$k=1$ to $K$}
        \item Generate a random initialization $\boldtheta_0^{(k)}$. 
        \item Execute the weight learning algorithm to get the weighting function $w^{(k)}=\mathbb{A}(\boldsymbol{x}, \boldtheta_0^{(k)})$. 
        \item Calculate discrete sample weights $\boldsymbol{W}^{(k)}=w^{(k)}(\boldsymbol{x})$. 
        \item Add $\boldsymbol{W}^{(k)}$ to $\tilde{\boldsymbol{W}}$. 
    \ENDFOR
    \STATE Average sets of sample weights in $\tilde{\boldsymbol{W}}$ for the ensemble result $\bar{\boldsymbol{W}}$. 
    \STATE {\bfseries return: $\bar{\boldsymbol{W}}$} 
\end{algorithmic}
\end{algorithm}

Since the learning processes of these sets of sample weights can be easily parallelized, SAWA exhibits a low time cost in contrast to SVI \citep{yu2023stable}, which incurs a high time cost due to its iterative framework that can hardly be parallelized. 
Meanwhile, this strategy does not require information from outcome labels \citep{yu2023stable} or environment labels \citep{shen2020stable2}, and can be flexibly incorporated into any existing independence-based sample reweighting methods, since the weight learning algorithm $\mathbb{A}$ in Algorithm \ref{alg:sawa} can be DWR, SRDO, SVI or any other ones. 
Next, we provide theoretical results from two perspectives. Detailed proofs can be referred to in Appendix. 

\subsubsection{Validity of averaged sample weights}
\label{sec:validity}
We prove that the average of possible solutions is also a valid solution for weight learning. 

\begin{proposition}
\label{prop:dwr}
For a stronger version of DWR that constrains both weighted covariance and weighted mean equal to zero, when $n>\frac{p(p+1)}{2}+1$, it will have infinite solutions if solvable. Furthermore, the solution space is a convex set. 
\end{proposition}

\begin{proposition}
\label{prop:srdo}
For SRDO, when using the LSIF loss $\mathbb{E}_{\tilde{P}}[-w(\boldx)]+\mathbb{E}_{P}[w(\boldx)^2/2]$ to directly learn the density ratio, i.e. the weighting function $w$, if restricting $w$ coming from the linear parameterized weighting function family $\mathcal{W}_{\rm lin}=\{w_{\boldtheta}(\boldx)=a(\boldx)^T \boldtheta+b(\boldx)\ | \ a:\mathcal{X}\mapsto \mathbb{R}^p,b:\mathcal{X}\mapsto \mathbb{R}\}$, then the minima constitute a convex set. 
\end{proposition}

For \Cref{prop:srdo}, the weighting function family $\mathcal{W}_{\rm lin}$ is rich because functions $a$ and $b$ can arbitrarily change and the dimension of $\boldtheta$ can be very high. In our implementation of SRDO, we use MLP-structured weighting functions. With proper assumptions, we can use Neural Tangent Kernel (NTK) approximation \citep{lee2019wide} to include wide MLPs into $\mathcal{W}_{\rm lin}$. 
As the possible solutions constitute a convex set for both DWR and SRDO, the average of multiple optimization results also belongs to the set, thus also a possible optimization outcome of the corresponding weight learning algorithm. So we confirm the validity and rationality of sample weight averaging. Note that other reweighting algorithms like SVI are based on DWR and SRDO, to which \Cref{prop:dwr} and \ref{prop:srdo} can also be applied. 

\subsubsection{Benefits of decreasing error of weight learning and model parameter estimation}
\label{sec:benefits}
Following the theory of bagging \citep{ghojogh2019theory} and existing theoretical analyses for model parameter averaging \citep{rame2022diverse}, we come up with the following proposition. 
\begin{proposition}    
Denote $w$ as some desired weighting function in $\mathcal{W}_{\perp}$. 
Denote $w^E$ as the expected weighting function outputted by a single weight learning procedure over the joint distribution $P^g$ of training data $\boldsymbol{x}$ and random initialization $\boldtheta_0$, calculated as $w^E(\boldx)=\expect{g\sim P^g}{w^g(\boldx)}$, where $g=(\boldsymbol{x}, \boldtheta_0)$. 
Denote $\bar{w}$ as the average of the $K$ learned weighting functions, calculated as $\bar{w}(\boldx)=\frac1K \sum_{k=1}^K w^{(k)}(\boldx)$, where $w^{(k)}=\mathbb{A}(g^{(k}))$, $\{g^{(k)}\}_{k=1}^K$ are identically sampled from $P^g$, and all pairs of elements in $\{g^{(k)}\}_{k=1}^K$ shares the same covariance. 
Then expected estimation error of the averaged weighting function over $P^{te}$ and $P^g$ can be decomposed into the following three parts:
\begin{equation} \label{eq:decomp}
\small
\begin{aligned}
&\expect{\left\{g^{(k)}\right\}_{k=1}^K}{\expect{\boldx\sim P^{te}}{(\bar{w}(\boldx)-w(\boldx))^2}}\\
=& \mathop{\mathbb{E}}\limits_{\boldx\sim P^{te}}\Bigg[\left(w^E(\boldx)-w(\boldx)\right)^2\\
+&\frac1K \expect{g^{(k)}}{\left(w^{(k)}(\boldx)-w^E(\boldx)\right)^2}\\
+&\frac{K-1}{K}\expect{g^{(l)},g^{(m)} 
\atop 
l\neq m}{\left(w^{(l)}(\boldx)-w^E(\boldx)\right)\left(w^{(m)}(\boldx)-w^E(\boldx)\right)}\Bigg]
\end{aligned}
\end{equation}
\label{prop:decomp}
\end{proposition}

The first term of the right-hand side is $(w^E(\boldx)-w(\boldx))^2$, the squared bias of weight learning, solely related to the weight learning algorithm itself. It remains constant irrespective of the averaging strategy. 
The second term can be interpreted as the variance of weight learning, inversely proportional to $K$. Therefore, when we apply SAWA, an increase of $K$ results in a reduction of this variance component. 
The third term characterizes the degree of diversity present in sample weights. It depicts the correlation between two distinct weighting functions. By enhancing the diversity among weighting functions used for averaging, this term can be mitigated. This can elucidate the superiority of averaging sample weights from different initializations, as compared with moving average from the same initialization, which is popular in current DG research \citep{cha2021swad,arpit2022ensemble}. This finding aligns with the conclusion drawn by~\citet{rame2022diverse}. Relevant empirical analyses are in \Cref{fig:sim-comp} of \Cref{sec:synthetic}.

Finally, following~\citet{xu2021stable}, we connect weight learning error with regression coefficient estimation error. 
\begin{proposition}
    Denote $\hat{\boldb}_{\bar{w}}$ as the model coefficient estimated by WLS using $\bar{w}$ with sample size $n$. 
    Denote $\boldb_w$ as the model coefficient estimated by WLS using some $w\in \mathcal{W}_{\perp}$ with infinite samples. 
    Denote $\Lambda_w$ as the smallest eigenvalue of the population-level weighted covariance matrix. 
    Then with mild assumptions, we have:
    \begin{equation}
        \left\|\hat{\boldb}_{\bar{w}}-\boldb_w\right\|\leq \frac{4\epsilon^2 M_w}{\left(\Lambda_w-\epsilon\sqrt{\expect{}{\|\boldx\|_2^4}}\right)^2}+O\left(\frac1n\right)
    \end{equation}
    where $\epsilon^2=\expect{\boldx\sim P^{tr}}{(\bar{w}(\boldx)-w(\boldx))^2}$ is the weight learning error, $M_w$ is a term only related to $w$. 
\label{prop:error}
\end{proposition}
\Cref{prop:error} reveals that as $n$ grows large enough, the dominated term in the bound of coefficient estimation error is positively related to the weight learning error. 
Notably, coefficients associated with unstable variables $\boldv$ in $\boldb_w$ are almost surely zero almost according to~\citet{xu2021stable}. 
Consequently, by refining the learning of sample weights, we can achieve improved estimations for the model coefficients on $\bolds$ and drive coefficients on $\boldv$ towards zero. Such refinement will lead to a stronger covariate-shift generalization ability and more stable prediction.

\begin{table*}[!ht]
\centering
\caption{Under linear settings, results with different strengths of collinearity in $\bolds$ and $\boldv$.
We highlight the better result w or w/o SAWA with the bold type, and underline the best result across all methods. Vary $n$ and $r$ while fixing $V_b=0.2*p$.}
\label{table:basic}
\resizebox{0.87\textwidth}{!}{%
\begin{tabular}{@{}ccccccccccccc@{}}
\toprule
\multicolumn{13}{c}{Scenario 1: Strong collinearity in $\bolds$ (Fixing $\rho_s=0.9, \rho_v=0.1, r=2.1$, varying $n$)}                                                                                                                                                                                                                                                                               \\ \midrule
\multicolumn{1}{c|}{$n$}       & \multicolumn{4}{c|}{$n=1000$}                                                                                  & \multicolumn{4}{c|}{$n=2000$}                                                                                  & \multicolumn{4}{c}{$n=15000$}                                                             \\ \midrule
\multicolumn{1}{c|}{Methods}   & $\beta$\_Error       & Mean\_Error          & Std\_Error           & \multicolumn{1}{c|}{Max\_Error}           & $\beta$\_Error       & Mean\_Error          & Std\_Error           & \multicolumn{1}{c|}{Max\_Error}           & $\beta$\_Error       & Mean\_Error          & Std\_Error           & Max\_Error           \\ \midrule
\multicolumn{1}{c|}{OLS}       & 0.895                & 0.393                & 0.057                & \multicolumn{1}{c|}{0.502}                & 0.815                & 0.388                & 0.048                & \multicolumn{1}{c|}{0.469}                & 0.765                & 0.382                & 0.045                & 0.461                \\
\multicolumn{1}{c|}{Ridge}     & 1.326                & 0.445                & 0.100                & \multicolumn{1}{c|}{0.612}                & 1.250                & 0.440                & 0.091                & \multicolumn{1}{c|}{0.572}                & 1.174                & 0.432                & 0.086                & 0.559                \\
\multicolumn{1}{c|}{Lasso}     & 1.761                & 0.494                & 0.138                & \multicolumn{1}{c|}{0.712}                & 1.657                & 0.483                & 0.124                & \multicolumn{1}{c|}{0.654}                & 1.603                & 0.472                & 0.116                & 0.635                \\
\multicolumn{1}{c|}{STG}       & -                    & 0.420                & 0.077                & \multicolumn{1}{c|}{0.557}                & -                    & 0.407                & 0.061                & \multicolumn{1}{c|}{0.503}                & -                    & 0.393                & 0.053                & 0.480                \\
\multicolumn{1}{c|}{DRO}       & 1.319                & 0.477                & 0.098                & \multicolumn{1}{c|}{0.626}                & 1.201                & 0.457                & 0.111                & \multicolumn{1}{c|}{0.582}                & 1.100                & 0.427                & 0.082                & 0.549                \\
\multicolumn{1}{c|}{JTT}       & 1.420                & 0.489                & 0.102                & \multicolumn{1}{c|}{0.656}                & 1.033                & 0.425                & 0.078                & \multicolumn{1}{c|}{0.519}                & 0.894                & 0.410                & 0.061                & 0.499                \\ \midrule
\multicolumn{1}{c|}{DWR}       & 1.432                & 0.600                & 0.054                & \multicolumn{1}{c|}{0.701}                & 0.693                & 0.464                & 0.038                & \multicolumn{1}{c|}{0.537}                & 0.614                & 0.391                & 0.038                & \textbf{0.441}       \\
\multicolumn{1}{c|}{DWR+SAWA}  & \textbf{1.308}       & \textbf{0.507}       & \textbf{0.039}       & \multicolumn{1}{c|}{\textbf{0.586}}       & {\ul \textbf{0.667}} & \textbf{0.407}       & \textbf{0.035}       & \multicolumn{1}{c|}{\textbf{0.472}}       & {\ul \textbf{0.557}} & \textbf{0.375}       & \textbf{0.026}       & 0.445                \\ \midrule
\multicolumn{1}{c|}{SRDO}      & 0.781                & 0.405                & \textbf{0.034}       & \multicolumn{1}{c|}{0.486}                & 0.961                & 0.410                & 0.028                & \multicolumn{1}{c|}{0.465}                & 0.644                & 0.428                & 0.030                & 0.485                \\
\multicolumn{1}{c|}{SRDO+SAWA} & {\ul \textbf{0.702}} & \textbf{0.392}       & 0.036                & \multicolumn{1}{c|}{\textbf{0.475}}       & \textbf{0.822}       & \textbf{0.407}       & \textbf{0.022}       & \multicolumn{1}{c|}{\textbf{0.445}}       & \textbf{0.577}       & \textbf{0.418}       & \textbf{0.030}       & \textbf{0.476}       \\ \midrule
\multicolumn{1}{c|}{SVI}       & -                    & 0.376                & 0.028                & \multicolumn{1}{c|}{0.441}                & -                    & 0.380                & 0.028                & \multicolumn{1}{c|}{0.429}                & -                    & 0.409                & 0.064                & 0.510                \\
\multicolumn{1}{c|}{SVI+SAWA}  & -                    & {\ul \textbf{0.359}} & {\ul \textbf{0.016}} & \multicolumn{1}{c|}{{\ul \textbf{0.400}}} & -                    & {\ul \textbf{0.376}} & {\ul \textbf{0.015}} & \multicolumn{1}{c|}{{\ul \textbf{0.353}}} & -                    & {\ul \textbf{0.340}} & {\ul \textbf{0.012}} & {\ul \textbf{0.362}} \\ \midrule
\multicolumn{1}{c|}{SVI'}      & -                    & 0.399                & 0.055                & \multicolumn{1}{c|}{0.506}                & -                    & 0.383                & 0.036                & \multicolumn{1}{c|}{0.448}                & -                    & 0.353                & 0.017                & 0.387                \\
\multicolumn{1}{c|}{SVI'+SAWA} & -                    & \textbf{0.373}       & \textbf{0.038}       & \multicolumn{1}{c|}{\textbf{0.453}}       & -                    & \textbf{0.384}       & \textbf{0.028}       & \multicolumn{1}{c|}{\textbf{0.435}}       & -                    & \textbf{0.341}       & \textbf{0.012}       & \textbf{0.363}       \\ \midrule
\multicolumn{13}{c}{Scenario 2: Strong collinearity in both $\bolds$ and $\boldv$ (Fixing $\rho_s=\rho_v=0.7$, varying $r$ and $n$)}                                                                                                                                                                                                                                                     \\ \midrule
\multicolumn{1}{c|}{$r$}       & \multicolumn{4}{c|}{$n=1000, r=2.1$}                                                                           & \multicolumn{4}{c|}{$n=1000, r=2.5$}                                                                           & \multicolumn{4}{c}{$n=2000, r=2.5$}                                                       \\ \midrule
\multicolumn{1}{c|}{Methods}   & $\beta$\_Error       & Mean\_Error          & Std\_Error           & \multicolumn{1}{c|}{Max\_Error}           & $\beta$\_Error       & Mean\_Error          & Std\_Error           & \multicolumn{1}{c|}{Max\_Error}           & $\beta$\_Error       & Mean\_Error          & Std\_Error           & Max\_Error           \\ \midrule
\multicolumn{1}{c|}{OLS}       & 1.414                & 0.532                & 0.158                & \multicolumn{1}{c|}{0.732}                & 1.661                & 0.602                & 0.218                & \multicolumn{1}{c|}{0.860}                & 1.622                & 0.602                & 0.217                & 0.877                \\
\multicolumn{1}{c|}{Ridge}     & 1.655                & 0.583                & 0.202                & \multicolumn{1}{c|}{0.828}                & 1.894                & 0.659                & 0.267                & \multicolumn{1}{c|}{0.969}                & 1.867                & 0.660                & 0.266                & 0.990                \\
\multicolumn{1}{c|}{Lasso}     & 1.911                & 0.653                & 0.261                & \multicolumn{1}{c|}{0.960}                & 2.283                & 0.783                & 0.370                & \multicolumn{1}{c|}{1.203}                & 2.296                & 0.787                & 0.372                & 1.231                \\
\multicolumn{1}{c|}{STG}       & -                    & 0.531                & 0.158                & \multicolumn{1}{c|}{0.729}                & -                    & 0.600                & 0.217                & \multicolumn{1}{c|}{0.857}                & -                    & 0.601                & 0.217                & 0.876                \\
\multicolumn{1}{c|}{DRO}       & 1.392                & 0.522                & 0.130                & \multicolumn{1}{c|}{0.653}                & 1.591                & 0.599                & 0.213                & \multicolumn{1}{c|}{0.849}                & 1.582                & 0.579                & 0.218                & 0.661                \\
\multicolumn{1}{c|}{JTT}       & 1.381                & 0.524                & 0.134                & \multicolumn{1}{c|}{0.698}                & 1.535                & 0.570                & 0.198                & \multicolumn{1}{c|}{0.794}                & 1.441                & 0.552                & 0.220                & 0.649                \\ \midrule
\multicolumn{1}{c|}{DWR}       & 1.342                & 0.519                & 0.055                & \multicolumn{1}{c|}{0.613}                & 2.259                & 0.861                & 0.146                & \multicolumn{1}{c|}{1.070}                & 1.364                & 0.682                & 0.040                & 0.772                \\
\multicolumn{1}{c|}{DWR+SAWA}  & \textbf{1.340}       & \textbf{0.516}       & {\ul \textbf{0.051}} & \multicolumn{1}{c|}{\textbf{0.606}}       & \textbf{2.244}       & \textbf{0.853}       & \textbf{0.127}       & \multicolumn{1}{c|}{\textbf{1.041}}       & \textbf{1.234}       & \textbf{0.563}       & \textbf{0.035}       & \textbf{0.643}       \\ \midrule
\multicolumn{1}{c|}{SRDO}      & 0.899                & 0.460                & 0.072                & \multicolumn{1}{c|}{0.568}                & 1.077                & 0.488                & 0.089                & \multicolumn{1}{c|}{0.618}                & 0.997                & 0.496                & 0.049                & 0.581                \\
\multicolumn{1}{c|}{SRDO+SAWA} & {\ul \textbf{0.868}} & {\ul \textbf{0.455}} & \textbf{0.057}       & \multicolumn{1}{c|}{{\ul \textbf{0.553}}} & {\ul \textbf{0.870}} & {\ul \textbf{0.465}} & {\ul \textbf{0.073}} & \multicolumn{1}{c|}{{\ul \textbf{0.578}}} & {\ul \textbf{0.992}} & \textbf{0.488}       & {\ul \textbf{0.026}} & \textbf{0.544}       \\ \midrule
\multicolumn{1}{c|}{SVI}       & -                    & 0.518                & 0.133                & \multicolumn{1}{c|}{0.690}                & -                    & 0.706                & 0.263                & \multicolumn{1}{c|}{0.991}                & -                    & 0.458                & \textbf{0.050}       & 0.543                \\
\multicolumn{1}{c|}{SVI+SAWA}  & -                    & \textbf{0.456}       & \textbf{0.067}       & \multicolumn{1}{c|}{\textbf{0.558}}       & -                    & \textbf{0.653}       & \textbf{0.222}       & \multicolumn{1}{c|}{\textbf{0.904}}       & -                    & {\ul \textbf{0.439}} & 0.056                & {\ul \textbf{0.533}} \\ \midrule
\multicolumn{1}{c|}{SVI'}      & -                    & 0.500                & 0.120                & \multicolumn{1}{c|}{0.657}                & -                    & 0.594                & 0.202                & \multicolumn{1}{c|}{0.835}                & -                    & 0.493                & 0.104                & 0.638                \\
\multicolumn{1}{c|}{SVI'+SAWA} & -                    & \textbf{0.472}       & \textbf{0.099}       & \multicolumn{1}{c|}{\textbf{0.608}}       & -                    & \textbf{0.534}       & \textbf{0.153}       & \multicolumn{1}{c|}{\textbf{0.728}}       & -                    & \textbf{0.490}       & \textbf{0.100}       & \textbf{0.631}       \\ \bottomrule
\end{tabular}%
}
\centering
\caption{Under nonlinear settings, results with varying $n$ and $r$. We highlight the better result w or w/o SAWA with the bold type, and underline the best result across all methods. Vary $n$ and $r$ while fixing $\rho_s=0.9, \rho_v=0.1, V_b=0.1*p$. }
\label{table:nonlinear}
\resizebox{0.81\textwidth}{!}{%
\begin{tabular}{@{}cccccccccc@{}}
\toprule
\multicolumn{10}{c}{Scenario 1: Varying sample size $n$ (Fixing $r=2.0$)}                                                                                                                                                                                                              \\ \midrule
\multicolumn{1}{c|}{$n$}       & \multicolumn{3}{c|}{$n$=15000}                                                          & \multicolumn{3}{c|}{$n=20000$}                                                          & \multicolumn{3}{c}{$n=25000$}                                      \\ \midrule
\multicolumn{1}{c|}{Methods}   & Mean\_Error          & Std\_Error           & \multicolumn{1}{c|}{Max\_Error}           & Mean\_Error          & Std\_Error           & \multicolumn{1}{c|}{Max\_Error}           & Mean\_Error          & Std\_Error           & Max\_Error           \\ \midrule
\multicolumn{1}{c|}{MLP}       & 0.221                & 0.080                & \multicolumn{1}{c|}{0.331}                & 0.262                & 0.113                & \multicolumn{1}{c|}{0.416}                & 0.249                & 0.104                & 0.389                \\
\multicolumn{1}{c|}{STG}       & 0.177                & 0.049                & \multicolumn{1}{c|}{0.243}                & 0.176                & 0.048                & \multicolumn{1}{c|}{0.241}                & 0.176                & 0.048                & 0.243                \\
\multicolumn{1}{c|}{DRO}       & 0.289                & 0.178                & \multicolumn{1}{c|}{0.402}                & 0.281                & 0.142                & \multicolumn{1}{c|}{0.477}                & 0.301                & 0.132                & 0.434                \\
\multicolumn{1}{c|}{JTT}       & 0.209                & 0.075                & \multicolumn{1}{c|}{0.298}                & 0.241                & 0.100                & \multicolumn{1}{c|}{0.295}                & 0.167                & 0.033                & 0.210                \\ \midrule
\multicolumn{1}{c|}{SRDO}      & 0.244                & 0.123                & \multicolumn{1}{c|}{0.380}                 & 0.288                & 0.133                & \multicolumn{1}{c|}{0.469}                & 0.231                & 0.090                & 0.373                \\
\multicolumn{1}{c|}{SRDO+SAWA} & \textbf{0.163}       & \textbf{0.031}       & \multicolumn{1}{c|}{\textbf{0.211}}       & \textbf{0.169}       & \textbf{0.037}       & \multicolumn{1}{c|}{\textbf{0.235}}       & \textbf{0.151}       & \textbf{0.026}       & \textbf{0.198}       \\ \midrule
\multicolumn{1}{c|}{SVI'}      & 0.130                & 0.002                & \multicolumn{1}{c|}{0.133}                & 0.125                & 0.001                & \multicolumn{1}{c|}{0.128}                & 0.126                & {\ul \textbf{0.002}} & 0.129                \\
\multicolumn{1}{c|}{SVI'+SAWA} & {\ul \textbf{0.126}} & {\ul \textbf{0.002}} & \multicolumn{1}{c|}{{\ul \textbf{0.130}}} & {\ul \textbf{0.122}} & {\ul \textbf{0.001}} & \multicolumn{1}{c|}{{\ul \textbf{0.126}}} & {\ul \textbf{0.109}} & 0.003                & {\ul \textbf{0.127}} \\ \midrule
\multicolumn{10}{c}{Scenario 2: Varying bias rate $r$ (Fixing $n=25000$)}                                                                                                                                                                                                              \\ \midrule
\multicolumn{1}{c|}{$r$}       & \multicolumn{3}{c|}{$r=1.8$}                                                            & \multicolumn{3}{c|}{$r=2.0$}                                                            & \multicolumn{3}{c}{$r=2.2$}                                        \\ \midrule
\multicolumn{1}{c|}{Methods}   & Mean\_Error          & Std\_Error           & \multicolumn{1}{c|}{Max\_Error}           & Mean\_Error          & Std\_Error           & \multicolumn{1}{c|}{Max\_Error}           & Mean\_Error          & Std\_Error           & Max\_Error           \\ \midrule
\multicolumn{1}{c|}{MLP}       & 0.188                & 0.051                & \multicolumn{1}{c|}{0.259}                & 0.249                & 0.104                & \multicolumn{1}{c|}{0.389}                & 0.498                & 0.312                & 0.901                \\
\multicolumn{1}{c|}{STG}       & 0.150                & 0.026                & \multicolumn{1}{c|}{0.186}                & 0.176                & 0.048                & \multicolumn{1}{c|}{0.243}                & 0.208                & 0.076                & 0.308                \\
\multicolumn{1}{c|}{DRO}       & 0.212                & 0.098                & \multicolumn{1}{c|}{0.345}                & 0.272                & 0.137                & \multicolumn{1}{c|}{0.422}                & 0.569                & 0.435                & 1.300                \\
\multicolumn{1}{c|}{JTT}       & 0.149                & 0.027                & \multicolumn{1}{c|}{0.195}                & 0.160                & 0.032                & \multicolumn{1}{c|}{0.256}                & 0.219                & 0.074                & 0.295                \\ \midrule
\multicolumn{1}{c|}{SRDO}      & 0.200                & 0.071                & \multicolumn{1}{c|}{0.297}                & 0.236                & 0.089                & \multicolumn{1}{c|}{0.353}                & 0.469                & 0.203                & 0.717                \\
\multicolumn{1}{c|}{SRDO+SAWA} & \textbf{0.144}       & \textbf{0.023}       & \multicolumn{1}{c|}{\textbf{0.182}}       & \textbf{0.154}       & \textbf{0.023}       & \multicolumn{1}{c|}{\textbf{0.210}}       & \textbf{0.201}       & \textbf{0.046}       & \textbf{0.272}       \\ \midrule
\multicolumn{1}{c|}{SVI'}      & 0.132                & 0.007                & \multicolumn{1}{c|}{0.144}                & 0.126                & 0.002                & \multicolumn{1}{c|}{0.129}                & 0.126                & 0.002                & 0.129                \\
\multicolumn{1}{c|}{SVI'+SAWA} & {\ul \textbf{0.120}} & {\ul \textbf{0.006}} & \multicolumn{1}{c|}{{\ul \textbf{0.135}}} & {\ul \textbf{0.121}} & {\ul \textbf{0.002}} & \multicolumn{1}{c|}{{\ul \textbf{0.126}}} & {\ul \textbf{0.121}} & {\ul \textbf{0.002}} & {\ul \textbf{0.125}} \\ \bottomrule
\end{tabular}%
}
\end{table*}

\begin{figure*}[t]
	\centering
	\subfigure[RMSE when varying the selection bias rate of test environments, fixing $n=1000$. ]  {
	\label{fig:linear}
	    \includegraphics[width=0.3\linewidth]{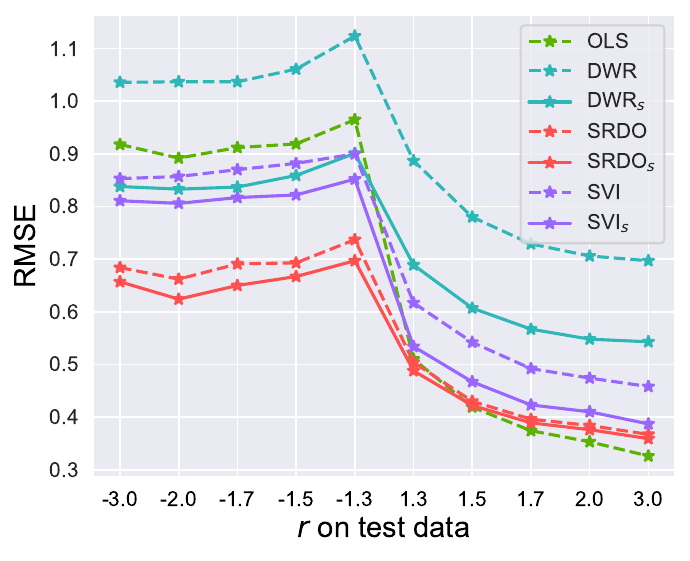}
	}
	\subfigure[Bias and variance of model parameter estimation for DWR w or w/o SAWA.] {
	\label{fig:bv}
	    \includegraphics[width=0.3\linewidth]{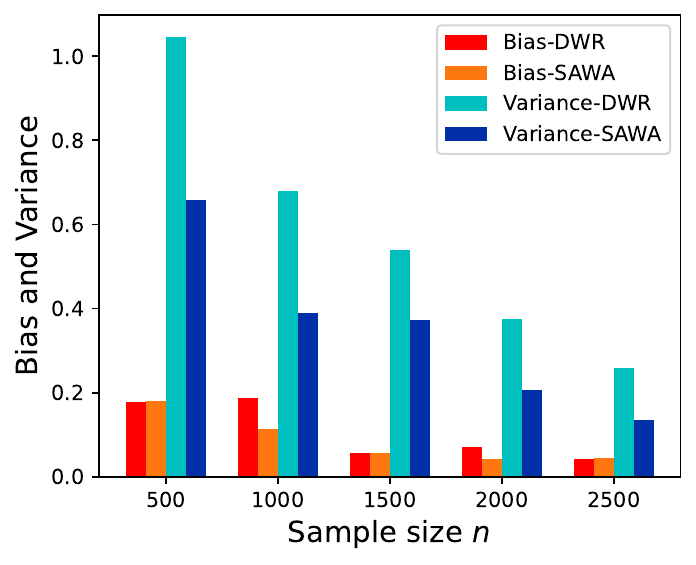}
	}
	\subfigure[RMSE when varying the number of averaged sets of sample weights.]   {
	\label{fig:ens}
	    \includegraphics[width=0.3\linewidth]{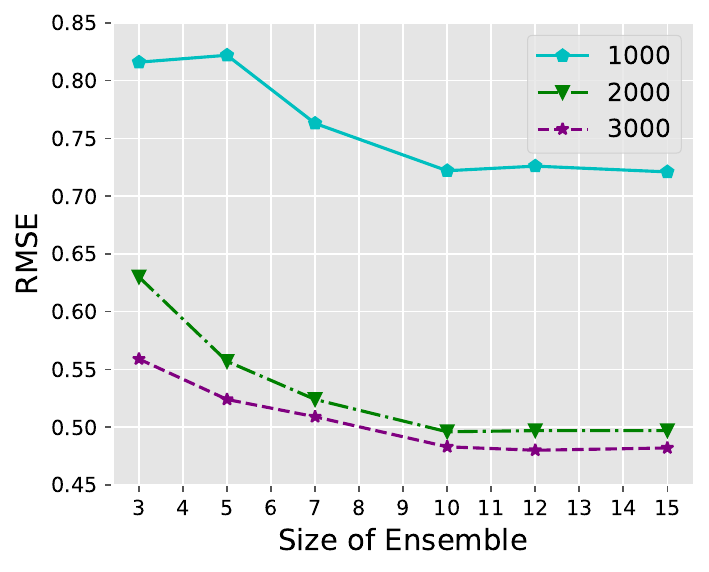}
	}
	\caption{Results on synthetic data when fixing $r_{train}=3.0,\ \rho_s=0.7,\ \rho_v=0.1$. 
    Subscript $_s$ represents combination with SAWA, drawn in solid lines while baselines are drawn in dashed lines. 
    In Figure \ref{fig:linear}, we can see that SAWA helps every sample reweighting method decrease the prediction error. 
    In Figure \ref{fig:bv}, we can see that SAWA greatly mitigates the variance for DWR while keeping the bias in a moderate range. 
    In Figure \ref{fig:ens}, we can see that the reduction of RMSE becomes marginal after the number of ensemble exceeds 10, so we set it as the constant value when we apply SAWA. 
    }
	\label{fig:basic}
\end{figure*}

\section{Experiments}
\label{sec:exp}

We have carried out comprehensive experiments on both synthetic and real-world datasets to demonstrate the superior performance of the proposed SAWA. We have also conducted abundant extra experiments to facilitate deeper and more thorough analyses. 

\subsection{Baselines}

Here we list the baseline methods we compare to. 
\begin{itemize}
    \item OLS: Minimizing the residual sum of squares (RSS). 
    \item Ridge~\citep{tikhonov1963solution}: Minimizing RSS and $\ell_2$ norm. 
    \item Lasso~\citep{tibshirani1996regression}: Minimizing RSS and $\ell_1$ norm. 
    \item STG~\citep{yamada2020feature}: Feature selection via a probablitic and continuous approximation of $\ell_0$ norm, exhibiting strong performance. 
    \item DRO~\citep{sinha2018certifying}: Searching for the worst-case distribution and optimizing for it. 
    \item JTT~\citep{liu2021just}: Training with ERM in the first round, and upweighting samples with high losses in the second round. 
    \item DWR~\citep{kuang2020stable}: Optimizing weighted covariance to learn sample weights and conduct WLS. 
    \item SRDO~\citep{shen2020stable}: Employing density ratio estimation to learn a weighting function and conduct WLS. 
    \item SVI~\citep{yu2023stable}: An iterative procedure combining sample reweighting and sparsity constraint.  
    \item SVI'~\citep{yu2023stable}: The nonlinear version of SVI. 
\end{itemize}
For linear settings, we exclude MLP. For nonlinear settings, we do not report linear algorithms (OLS, Ridge, Lasso) or reweighting algorithms of linear decorrelation (DWR and SVI). 

\subsection{Evaluation Metrics}

We evaluate our strategy across multiple environments to assess its covariate-shift generalization ability. The metrics adopted are listed below:
\begin{itemize}
    \item $\beta\_{\rm Error}=\|\hat{\boldsymbol{\beta}}-\boldsymbol{\beta}\|_1$; 
    \item ${\rm Mean\_Error} = \frac{1}{|\varepsilon_{te}|}\sum_{e\in \varepsilon_{te}}\mathcal{L}^e$;
    \item ${\rm Std\_Error} = \sqrt{\frac{1}{|\varepsilon_{te}|-1} \sum_{e\in \varepsilon_{te}}(\mathcal{L}^e-Mean\_Error)^2}$;
    \item ${\rm Max\_Error} = \max_{e\in \varepsilon_{te}}\mathcal{L}^e$;
\end{itemize}
Here $\varepsilon_{te}$ refers to test environments. $\mathcal{L}^e$ is empirical error in the environment $e$. 
Among the metrics, $\beta$\_Error is only adopted for linear settings. 
For STG, SVI, and SVI', they conduct a feature selection procedure and then apply OLS on the selected features, so we do not report $\beta$\_Error for them.

\begin{figure*}[t]
	\centering
	\subfigure[Comparing sample weights corresponding to MA and SAWA.] {
	\label{fig:dist-comp}
	    \includegraphics[width=0.3\linewidth]{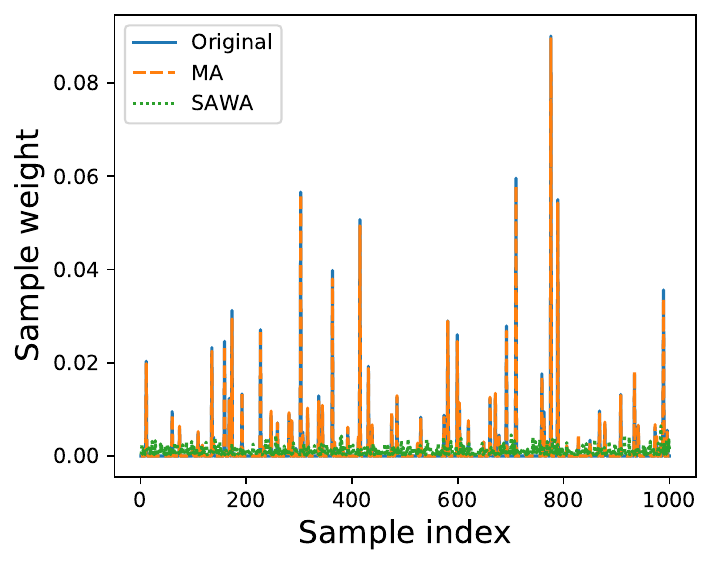}
	}
	\subfigure[Similarity of sample weights generated by MA and SAWA.] {
	\label{fig:sim-comp}
	    \includegraphics[width=0.3\linewidth]{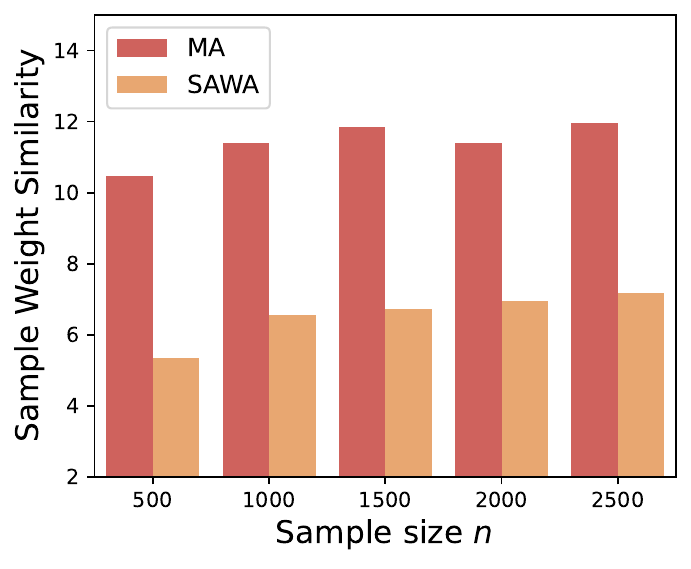}
	}
	\subfigure[Comparing prediction error of MA, CA and SAWA.] {
	\label{fig:avg-comp}
	    \includegraphics[width=0.3\linewidth]{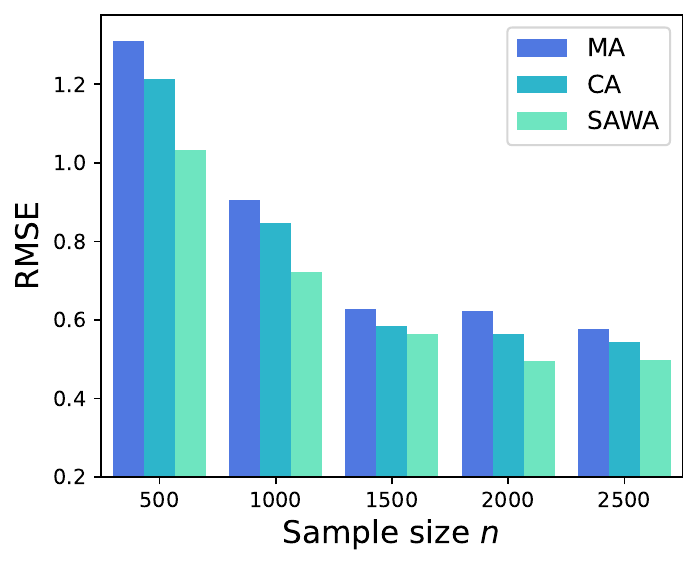}
	}
	\caption{Comparison with moving average (MA) and coefficient average (CA) when fixing $r_{train}=3.0,\ \rho_s=0.7,\ \rho_v=0.1$. 
    In Figure \ref{fig:dist-comp}, SAWA generates a very different set of sample weights from the original one, while MA nearly overlaps with the original one. 
    In Figure \ref{fig:sim-comp}, there is a lower similarity among the sets of sample weights that SAWA generates than MA. 
    In Figure \ref{fig:avg-comp}, SAWA achieves lower prediction error than other parameter averaging strategies. 
    }
	\label{fig:comp}
\end{figure*}

\subsection{Experiments on Synthetic Data}
\label{sec:synthetic}

\subsubsection{Data generation}
Before simulating different environments via selection bias, we first generate covariates $\boldx=\{\bolds, \boldv\}$ from $N(\boldsymbol{0}, \boldsigma)$, where $\boldsigma={\rm Diag}\left( \boldsigma^{(S)}, \boldsigma^{(V)} \right)$, assuming block structure. 
For $\boldsigma^{(S)} \in \mathbb{R}^{p_s \times p_s}$, we have: $\boldsigma^{(S)}_{jk}=\rho_s$ for $j \neq k$ and $\boldsigma^{(S)}_{jk}=1$ for $j=k$. 
For $\boldsigma^{(V)} \in \mathbb{R}^{p_v \times p_v}$, we have: $\boldsigma^{(V)}_{jk}=\rho_v$ for $j \neq k$ and $\boldsigma^{(V)}_{jk}=1$ for $j=k$. 
Thus we can control the strength of collinearity in stable variables $\bolds$ and unstable variables $\boldv$. 
Then we generate the outcome $Y$. 
For linear settings, we add a polynomial term to the linear term which still dominates the data generation:  
$
    Y=f(\boldsymbol{S})+\epsilon=[\boldsymbol{S}, \boldsymbol{V}]\cdot [\boldsymbol{\beta}_s, \boldsymbol{\beta}_v]^T + \boldsymbol{S}_{\cdot, 1}\boldsymbol{S}_{\cdot, 2}\boldsymbol{S}_{\cdot, 3} + \epsilon
$ and later use linear model to fit the data. 
For nonlinear settings, we generate data in a totally nonlinear way. We employ random initialized MLP as the data generation function: 
$
    Y=f(\boldsymbol{S})+\epsilon = MLP(\boldsymbol{S})+\epsilon 
$ and later use MLP with smaller capacity to fit the data. 
Note that a certain degree of model misspecification is needed, otherwise the model will be good enough to capture the stable variables already. 

To create covariate shift, we try to simulate various environments by changing $P(\boldv_b|\bolds)$, where $\boldv_b\in \boldv$. In this way, we can generate spurious correlations in $P(Y|\boldv)$ through a process of data selection following~\citet{kuang2020stable}: 
Given a bias rate $r\in[-3, -1)\cup(1,3]$, we select each sample with a probability of $Pr=\Pi_{V_i \in \boldsymbol{V}_b}|r|^{-5D_i}$, where $D_i=|f(\boldsymbol{S})-{\rm sign}(r)V_i|$ and ${\rm sign}$ represents the sign function.
We can find that $r>1$ corresponds to a positive correlation between $Y$ and $\boldsymbol{V}_b$, and $r<-1$ refers to a negative correlation between $Y$ and $\boldsymbol{V}_b$. 
A larger value of $|r|$ implies a stronger correlation between $\boldsymbol{V}_b$ and $Y$. By varying the value of the bias rate $r$, we can simulate different environments.

\subsubsection{Experimental settings}
In our experiments, for linear settings, we simulate two scenarios: strong collinearity in $\bolds$ but weak collinearity in $\boldv$, and strong collinearity in both $\bolds$ and $\boldv$. The second one is more challenging, and is beyond the structural assumption made in~\citep{yu2023stable}. 
For nonlinear settings, we assume strong collinearity in $\bolds$ and weak collinearity in $\boldv$. 
For the hyperparameter introduced by our strategy, i.e. the number of averaged sets of sample weights, we fix it as $10$, which will be analyzed and explained later. 
Other hyperparameters are selected according to the performance on validation data which is split from training data. 
In each specific experimental setting, we employ a single environment corresponding to the bias rate $r_{train}$ for training, and test on multiple environments with $r_{test}$ ranging in $[-3, -1)\cup(1,3]$. 
We run all the experiments $10$ times and report the averaged outcome of these 10 times unless otherwise specified. 
More details regarding data generation and synthetic experimental settings are included in Appendix. We leave analyses of time and memory in Appendix as well.

\subsubsection{Basic results} \Cref{table:basic} and \ref{table:nonlinear} display the estimation errors and prediction performance of baselines and our strategy. For each sample reweighting scheme and its combination with SAWA, superior results (indicated by lower values) are highlighted in bold typeface. The optimal result among all methods is underlined for each setting. 
As we can see, the integration with SAWA yields improvements when combined with four different independence-based sample reweighting algorithms, enhancing model parameter estimation and the average OOD prediction performance across all settings. 
Meanwhile, SAWA also reduces Std\_Error and Max\_Error in most settings, implying that it improves the stability of test performance. 
Such improvements are also witnessed in nonlinear settings. 
We can refer to Figure~\ref{fig:linear} as well, where improvements are made across all test environments when paired with established reweighting techniques (solid lines), especially when $r_{test}<0$, signifying a reversed correlation compared with $r_{train}$. 
It is worth noting that in linear settings SAWA brings improvement in both scenarios, where the scenario of strong collinearity in both $\bolds$ and $\boldv$ has never been well addressed before. 
Overall, the flexibility of seamlessly integrating with any reweighting method, coupled with improvements observed across all settings, collectively demonstrates the efficacy and universality of our strategy against covariate shift.

\subsubsection{Bias-variance analysis} Early reweighting methods suffer from variance inflation due to inefficient utilization of samples, an inherent vulnerability of the reweighting operation itself. 
In \Cref{fig:bv}, it is evident that after applying SAWA, variance of model parameter estimation diminishes sharply while the bias increases by a tiny margin or decreases, leading to better bias-variance trade-off. 

\subsubsection{Analysis for the size of ensemble} The size of sample weights ensemble is a hyperparameter in our strategy. We conduct experiments to assess its influence on the performance. From Figure \ref{fig:ens}, we can see that after exceeding $10$, the reduction in RMSE becomes negligible. Thus we fix it to a constant value of $10$ across all conducted experiments.

\begin{figure*}[!t]
	\centering
	\subfigure[Prediction error for house price.] {
	\label{fig:house}
	    \includegraphics[width=0.3\linewidth]{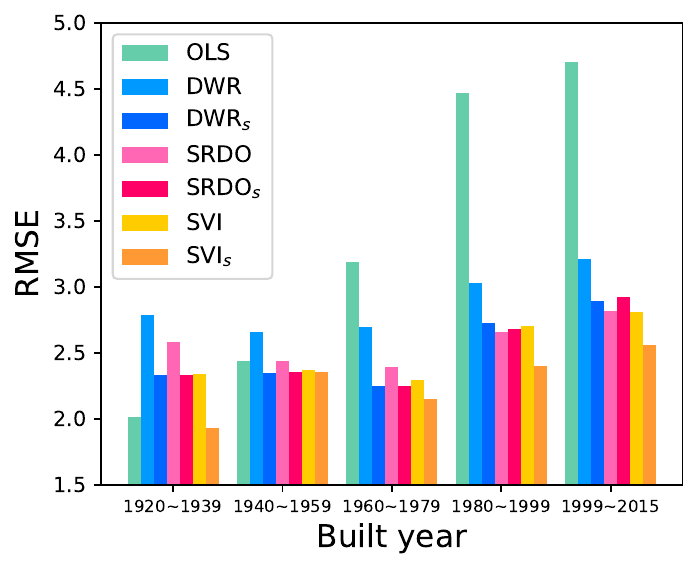}
	}
	\subfigure[Classification error for Adult.] {
	\label{fig:adult}
	    \includegraphics[width=0.3\linewidth]{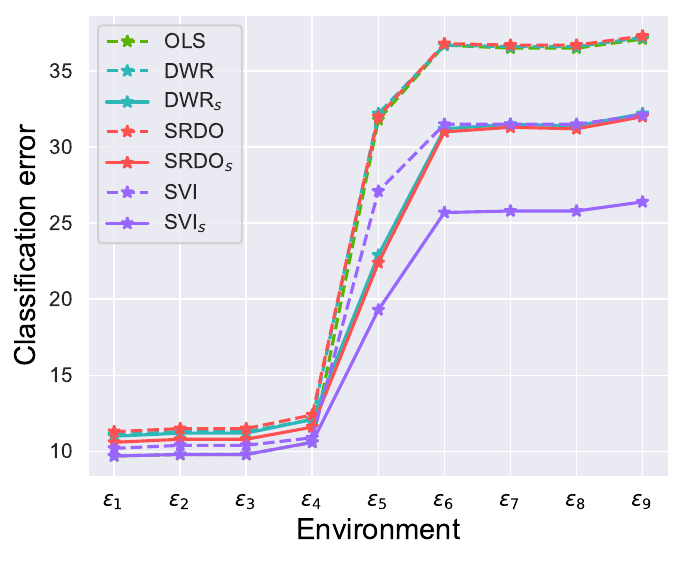}
	}
	\subfigure[Classification error for C-MNIST] {
	\label{fig:cmnist}
	    \includegraphics[width=0.3\linewidth]{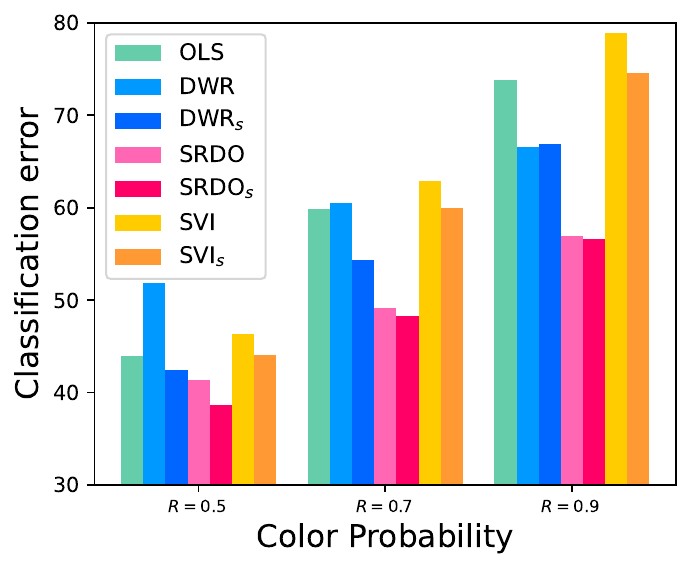}
	}
 	\subfigure[Classification error for ACS PUMS data] {
	\label{fig:acs}
	    \includegraphics[width=\linewidth]{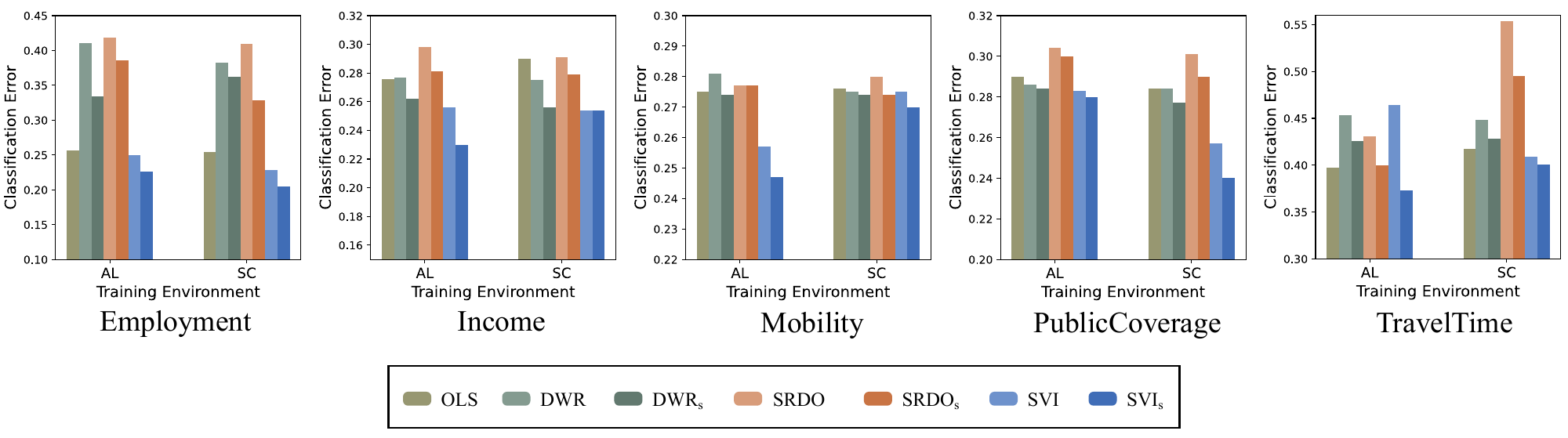}
	}
	\caption{Results of experiments on real-world data. The subscript $_s$ represents a combination with SAWA. We use similar colors for a certain reweighting method w or w/o SAWA (darker or lighter). 
    For the convenience of plotting, we only plot 7 methods. The detailed results of other methods are in Appendix. We can see that after being combined with SAWA, all sample reweighting methods gain a decrease in prediction error against distribution shifts. 
 }
 \label{fig:real}
\end{figure*}

\subsubsection{Comparison with other averaging strategies}
Previous OOD works employ moving average (MA) \citep{arpit2022ensemble} and model parameter averaging from multiple runs (DiWA) \citep{rame2022diverse}. The latter averages model parameters learned with the same initialization but different hyperparameters, of which a direct adaptation to our linear setting is model coefficient averaging (CA). 
In \Cref{fig:dist-comp}, we plot the sample-wise distribution of sample weights of MA and SAWA along with the original weights. 
The result of MA almost overlaps with the original one, while that of SAWA exhibits remarkable differences. 
In \Cref{fig:sim-comp}, we plot the similarity of sample weights generated by MA and SAWA. Here we adopt the logarithmically transformed result of the third term in \Cref{eq:decomp} as the metric, which describes the correlation of different weighting functions. 
As observed, the similarity metric of SAWA is much lower. Combining these findings, we confirm that SAWA indeed produces significantly more diverse sets of sample weights. This conforms to theoretical analyses of \Cref{eq:decomp}, where the third term is more greatly decreased by SAWA than MA. 
As for CA, it also assembles results from multiple runs, but its estimation lacks theoretical properties for validity, while the sample weights acquired through SAWA can be guaranteed by \Cref{prop:dwr} and \ref{prop:srdo} to be valid and reasonable. Moreover, \Cref{fig:avg-comp} reveals that despite CA improving upon MA, it still falls behind SAWA by a large margin.

\subsection{Experiments on Real-World Data}
\label{sec:real}

We apply our strategy to multiple tabular data tasks and an image recognition task, following previous works \citep{arjovsky2019invariant, shen2020stable, liu2021heterogeneous}. 
Detailed information of the datasets are left in Appendix. 

\subsubsection{House price prediction}
It is a regression task predicting house prices based on attributes like the number of bedrooms or bathrooms. 
We partition the dataset into six distinct periods, each spanning two decades, treated as 6 environments. To examine the covariate-shift generalization ability, we train on the first period and test on the other five. 
As depicted in \Cref{fig:house}, in most cases SAWA brings improvement to current reweighting methods. This demonstrates the benefits of SAWA in real-world tasks. 

\subsubsection{People income prediction}
\label{exp:adult}
It is binary classification for predicting if the annual income of an adult surpasses 50k based on UCI Adult \citep{kohavi1996scaling}. We split the dataset into ten environments according to combination of race and sex. We train on the environment of (White, Female) and test on the others. From \Cref{fig:adult}, when testing on the last five environments, characterized by a male sex attribute, we observe a large improvement after applying SAWA. 

\subsubsection{Colored MNIST}
Following~\citet{arjovsky2019invariant}, we convert the labels of MNIST into binary format: assigning $0$ to $0\sim 4$, and $1$ to $5\sim 9$. Then we assign a color id of each image identical to the digit label but with a probability of $R$ to flip it. We also introduce label noise by flipping the digit label with a probability $C$. 
We set $C=0.25$ across all environments, and set $R=0.25$ for the training environment. For test environments, we generate via $R\in\{0.5,0.7,0.9\}$ for a reverse correlation compared with training data. From Figure \ref{fig:cmnist}, SAWA still decreases the error when applied to existing methods. 
This reveals the potential of our strategy for extension to vision tasks.

\subsubsection{ACS PUMS data} \citet{ding2021retiring} provide five binary classification tasks from US-wide ACS PUMS data in terms of employment, income, mobility, public coverage, and travel time. For each task, we randomly select 2 states 'AL' and 'SC' to establish 2 settings. For each setting, We treat 'AL' or 'SC' as the training environment and use the other 49 states as the test environments. Although ACS PUMS is a superset of the UCI Adult dataset used in Sec \ref{exp:adult}, we generate different distribution shifts for them. For UCI Adult, the shift comes from race and sex. For ACS PUMS, the shift comes from states, i.e. geographical locations. 
The results are in Figure \ref{fig:acs}. We observe increases in the test performance for each sample reweighting algorithm across all tasks after applying SAWA, which demonstrates the benefits of SAWA in terms of stable prediction.

\section{Conclusion}

In this paper, we propose a sample weight averaging strategy to address variance inflation of previous independence-based sample reweighting algorithms. 
We prove its validity and benefits with theoretical analyses. 
Extensive experiments across synthetic and multiple real-world datasets demonstrate its superiority in mitigating variance inflation and improving covariate-shift generalization.

\bibliographystyle{ACM-Reference-Format}
\bibliography{sample-base}

\end{document}